\documentclass[fleqn,10pt]{wlscirep}
\usepackage[utf8]{inputenc}
\usepackage[T1]{fontenc}
\usepackage{bm}
\usepackage{float}
\usepackage{multirow}
\title{Physics-informed self-supervised learning for predictive modeling of coronary artery digital twins}

\author[1,2,6,*]{Xiaowu Sun}
\author[3]{Thabo Mahendiran}
\author[1,2]{Ortal Senouf}
\author[3]{Denise Auberson}
\author[3,4]{Bernard De Bruyne}
\author[3]{Stephane Fournier}
\author[3]{Olivier Muller}
\author[2,5]{Pascal Frossard}
\author[1,5]{Emmanuel Abbé}
\author[2,5,*]{Dorina Thanou}

\affil[1]{Chair of Mathematical Data Science, EPFL, Lausanne, Switzerland}
\affil[2]{LTS4 laboratory, EPFL, Lausanne, Switzerland}
\affil[3]{Cardiology Department, Lausanne University Center Hospital, Lausanne, Switzerland}
\affil[4]{OLV Hospital, Aalst, Belgium}
\affil[5]{AI Center, EPFL, Lausanne, Switzerland}
\affil[6]{School of AI and Advanced Computing, Xi'an Jiaotong-Liverpool University, China}
\affil[*]{e-mail: xiaowu.sun@epfl.ch, dorina.thanou@epfl.ch}

\begin{abstract}
Cardiovascular disease is the leading global cause of mortality, with coronary artery disease (CAD) representing its most prevalent subtype, highlighting the urgent need for early and accurate risk prediction. Recent advances in imaging enable the reconstruction of 3D coronary artery digital twins, which capture anatomical detail beyond conventional 2D imaging and offer new opportunities for personalized assessment. However, analyzing digital twins typically relies on computational fluid dynamics (CFD) to extract hemodynamics, limiting scalability due to their high computational cost. Meanwhile, data-driven models are constrained by the limited availability of labeled 3D datasets and lack of built-in physiological priors. 
To address these challenges, we present PINS-CAD, a physics-informed self-supervised learning framework that pre-trains graph neural networks on a synthetic dataset of 200,000 coronary digital twins to predict pressure and flow distributions. Pretraining is guided by constraints from 1D Navier–Stokes equations and pressure-drop laws, enabling the model to learn physiologically informed hemodynamic representations without requiring CFD simulations or labeled flow data. Fine-tuned on clinical data from 635 patients in the multicenter FAME2 study, PINS-CAD achieves an AUC of 0.73 for predicting future cardiovascular events, outperforming clinical risk metrics and purely data-driven baselines. These results indicate that physics-informed pretraining enhances sample efficiency, enabling the model to extract physiologically meaningful representations from unlabeled data and reducing reliance on large CFD labeled cohorts.
Beyond predictive performance, PINS-CAD generates spatially resolved pressure and fractional flow reserve curves along coronary centerlines, offering physiologically interpretable biomarkers aligned with established pathophysiology. By embedding physical priors into geometric deep learning, PINS-CAD transforms routine angiography into a simulation-free, physiology-aware digital twins framework, opening the door to large-scaled, personalized models for preventive cardiology.
\end{abstract}

\begin{document}

\flushbottom
\maketitle

\thispagestyle{empty}

\section*{Introduction}
Cardiovascular diseases (CVDs) remain the leading cause of global mortality with coronary artery disease (CAD) representing the most prevalent subtype~\cite{timmis2024european}. 
Invasive coronary angiography (ICA) represents the gold standard test for CAD diagnosis. It permits the visualization of the coronary arteries and the anatomical evaluation of stenosis severity, such the quantification of diameter stenosis (DS) i.e. the percentage reduction in luminal diameter ~\cite{knuuti20202019}. However, anatomical narrowing does not always correspond to physiological impairment. Fractional flow reserve (FFR), the gold standard hemodynamic measure of lesion severity~\cite{barbato2016prospective}, provides complementary information on the functional impact of a stenosis, with current guidelines recommending the revascularisation of lesions with an FFR $\le$ 0.8~\cite{knuuti20202019}. However, whilst the combination of an angiographic anatomical assessment with FFR permits the risk stratification of coronary stenoses, a significant proportion of such patients with intermediate stenoses with an $\text{FFR} > 0.8$ go on to experience adverse events during long-term follow-up~\cite{de2014fractional, xaplanteris2018five}. The limitations of this current approach, together with the invasive nature, procedural risk, and cost of FFR measurements, highlight the need for personalized, non-invasive predictive frameworks that capture the complex interplay between coronary anatomy and physiology.

Recently, three-dimensional (3D) coronary artery digital twins, represented as point clouds, have emerged as a promising approach to patient-specific cardiovascular modeling~\cite{coorey2022health}. Unlike conventional ICA-derived features, which are often restricted to local descriptors such as diameter stenosis,  3D digital twins capture detailed patient-specific vascular geometry with full spatial context\cite{beetz2024modeling,yang2024pointchd,beetz2023multi}. They preserve continuous vessel topology and spatial relationships, enabling the integration of both local morphological and global anatomical features\cite{qiao2025personalized,lyu2025personalized}.
The geometric diversity and anatomical variations in these digital twins provide a natural source of lesion augmentation that enhances model generalization and enables personalized assessment of coronary physiology and risk. Moreover, building upon this geometric foundation, digital twins further enable comprehensive hemodynamic analysis, providing richer, patient-specific quantification of flow parameters including pressure, velocity and wall shear stress (WSS) than conventional FFR. By bridging detailed anatomical representation with functional hemodynamics, 3D digital twins provide a unified framework for physiological characterization, data-driven biomarker discovery, and predictive modeling for personalized risk stratification.

Computational fluid dynamics (CFD) can be performed on 3D digital twins to quantify flow parameters, providing insights into lesion-specific hemodynamics for clinical assessment~\cite{thangaraj2024cardiovascular}. Compared with routine clinical practice, where these physiological estimates rely on invasive FFR measurements, adding procedural risk, cost and time, 3D CFD offers comparable diagnostic insights in a non-invasive and scalable way, albeit whilst being computationally intensive. Reduced-order physical models provide a means of increasing efficiency while retaining essential hemodynamic principles~\cite{pegolotti2024learning,willems2025probabilistic}. Among them, the one-dimensional (1D) Navier–Stokes equations offer a well-established approximation of pulsatile blood flow along the vessel centerline, enforcing conservation of mass and momentum under physiological viscosity and wall compliance~\cite{pradhan2024one,liu2024comprehensive,pfaller2021periodicity}. This formulation has been extensively validated as a fast and reliable surrogate to 3D CFD for estimating hemodynamic analysis in coronary arteries~\cite{hu2023novel}. Complementary to this, the pressure-drop law provides an analytical means to estimate pressure losses along the arterial segment, arising from both viscous friction and inertial effects due to changes in vessel geometry and flow conditions~\cite{mirramezani2020distributed}. It quantitatively links geometric features, such as lumen area reduction, curvature, and lesion length, to the energy dissipation that drives local pressure gradients. These formulations capture the dominant mechanisms of coronary hemodynamics at greatly reduced cost, forming a physical foundation for data-driven modeling. Recent advancements in deep learning, including PointNet-based architecture~\cite{qi2017pointnet} and graph neural networks (GNNs)~\cite{kipf2016semi}, have been developed to non-invasively estimate functional hemodynamic biomarkers such as FFR\cite{fossan2021machine,zhang2024constraint} and WSS\cite{ferdian2022wssnet,garrido2022machine}, with recent methods incorporating physical constraints directly into GNN loss functions for improved physiological consistency~\cite{xie2023conditional,suk2024mesh}. Nevertheless, most methods remain supervised, relying on CFD-generated labels that are costly and time-consuming to obtain. As a result, they mainly focus on reproducing known biomarkers rather than discovering new hemodynamic signatures relevant to clinical decision-making. Therefore, integrating physical foundation with self-supervised learning frameworks represents a promising path toward scalable, non-invasive, and CFD-free prediction of patient-specific coronary physiology.

In this work, we propose \textbf{PINS-CAD}, a \textbf{P}hysics-\textbf{In}formed, \textbf{S}elf-supervised learning framework for predictive modeling of \textbf{C}oronary \textbf{A}rtery \textbf{D}igital twins. Recognizing that hemodynamic features are vital for assessing lesion severity and cardiovascular risk, yet their estimation through CFD or invasive FFR remains resource-intensive, PINS-CAD integrates cardiovascular flow physics into a self-supervised pretraining stage. This enables models to learn physiologically meaningful hemodynamic representations from large-scale unlabeled coronary geometries without CFD supervision.
By coupling physics-informed learning guided by hemodynamic principles with downstream clinical fine-tuning, PINS-CAD bridges the gap between geometric modeling and clinical outcome prediction, offering a scalable and interpretable solution for non-invasive risk assessment in coronary artery disease. 
Specifically, we reconstruct 3D coronary artery digital twins from paired ICA images acquired from two approximately orthogonal views, which provide sufficient angular separation to resolve depth ambiguity and accurately recover the spatial geometry of the coronary artery(Fig.\ref{fig:framework}a). To increase the training dataset diversity, we generate 200,000 synthetic coronary artery digital twins using A\textsuperscript{3}M, an anatomy-aware augmentation method that recombines centerline and radius profiles from different real anatomies while introducing controlled perturbations of curvature, torsion, and vessel radius to mimic physiological variability (Fig.~\ref{fig:framework}b). These anatomically diverse digital twins provide a foundation for learning hemodynamic behavior directly from geometry, enabling the prediction of key quantities such as pressure and velocities without relying on computationally intensive CFD simulations. To predict pressure and velocities from the digital twins, each digital twin is represented as a graph where boundary points sampled from the artery surface serve as nodes and edges are defined based on spatial proximity to capture local geometric and morphological patterns. A GNN-based~\cite{hamilton2017inductive, kipf2016semi} backbone is trained on these graphs to extract geometric and morphological features to predict pressure and velocities along the artery centerline. To enable centerline-level predictions, node features extracted from the boundary graph are aggregated into the centerline. Training is guided by physics-based loss functions derived from the 1D Navier–Stokes equations and pressure drop laws (Fig.\ref{fig:framework}c), ensuring that the learned representations remain consistent with cardiovascular fluid dynamics. The pretrained GNN is then fine-tuned on real digital twins to perform downstream clinical tasks such as cardiovascular event prediction (Fig.\ref{fig:framework}d). By embedding fundamental physical principles into self-supervised pretraining, PINS-CAD ensures physiological consistency, reduces dependency on labeled CFD data, and transforms anatomical digital twins into predictive tools for preventive and personalized clinical decision-making.

\begin{figure}[H]
    \centering
    \includegraphics[width=1\linewidth]{figs/framework.png}
    \caption{
    Overview of the proposed PINS-CAD, physics-informed self-supervised learning framework for predictive modeling of coronary artery digital twins. The framework consists of four stages. 
    \textbf{a:} Image-to-digital twin construction. Paired ICA images from two views are used to reconstruct a 3D anatomical model of the coronary artery, referred to as a real digital twin with pointwise correspondence. 
    \textbf{b:} A\textsuperscript{3}M: Anatomy-aware augmentation module for synthetic digital twin generation. Two real digital twins are randomly selected: one contributes the artery centerline, the other provides the radius profile. The centerline undergoes geometric augmentation to simulate anatomical variation. A new synthetic digital twin is then generated by sweeping the radius along the augmented centerline to define its boundary. 
    \textbf{c:} Physics-informed self-supervised pretraining. A large dataset of 200{,}000 synthetic digital twins is used to pretrain a graph neural network (GNN). Artery graphs are constructed and the GNN is trained to predict pressure and velocity along the centerline, guided by physical constraints based on the 1D Navier–Stokes equations and pressure drop consistency. \textbf{d:} Fine-tuning for downstream tasks. The pretrained GNN is fine-tuned on real digital twins to predict future cardiovascular events using learned hemodynamic features.
    }
    \label{fig:framework}
\end{figure}

\section*{Results}
\subsection*{Dataset}
The Fractional Flow Reserve Versus Angiography in Multivessel Evaluation 2 RCT (FAME2) dataset~\cite{de2012fractional} is used in this study to construct coronary artery digital twins and evaluate the proposed framework. FAME2 includes 563 patients with stable CAD from 28 centers across Europe and North America. Each patient has at least one lesion in the major coronary arteries. For each patient, two rotational projections are performed to capture the arteries from multiple views, ranging from 1 to 9 views per artery. From the original angiographic videos, 1748 high-quality 2D images with clearly visible arteries are selected. To ensure reliable 3D digital twins reconstruction, only arteries imaged with at least two different views are used for digital twin construction, resulting in a total of 914 coronary artery digital twins. Among these, 634 contain at least one annotated lesion, while the remaining 281 are non-lesion cases. The 634 lesion digital twins are divided into five folds for cross-validation. In each fold, the training subset is combined with the 281 non-lesion digital twins to generate 40,000 synthetic digital twins using geometric perturbations and randomized anatomical variations. This results in a total of 200,000 synthetic samples across all five folds, which are used for self-supervised pretraining of the proposed framework. Downstream tasks are performed exclusively on the 634 lesion digital twins, as only lesions are clinically associated with future cardiovascular events. Among these, 176 correspond to positive cases with cardiovascular events, while the remaining 458 are negative, indicating the class imbalance that makes the prediction task more challenging and clinically realistic. The primary endpoint for future cardiovascular events is a composite of cardiac death, MI, and both urgent and non-urgent revascularization. Arteries and lesions segmentation annotations on the 2D ICA images are provided by the cardiology team at Lausanne University Hospital (CHUV), Switzerland, and are subsequently projected onto the corresponding 3D digital twins. Additionally, clinical information, including FFR and DS measurements at the lesion site, is collected for each case. Further details are provided in Table~\ref{tab:clinical_summary}.

\subsection*{PINS-CAD enables prediction of future cardiovascular events}
\begin{figure}[htbp]
    \centering
    \includegraphics[width=0.9\linewidth]{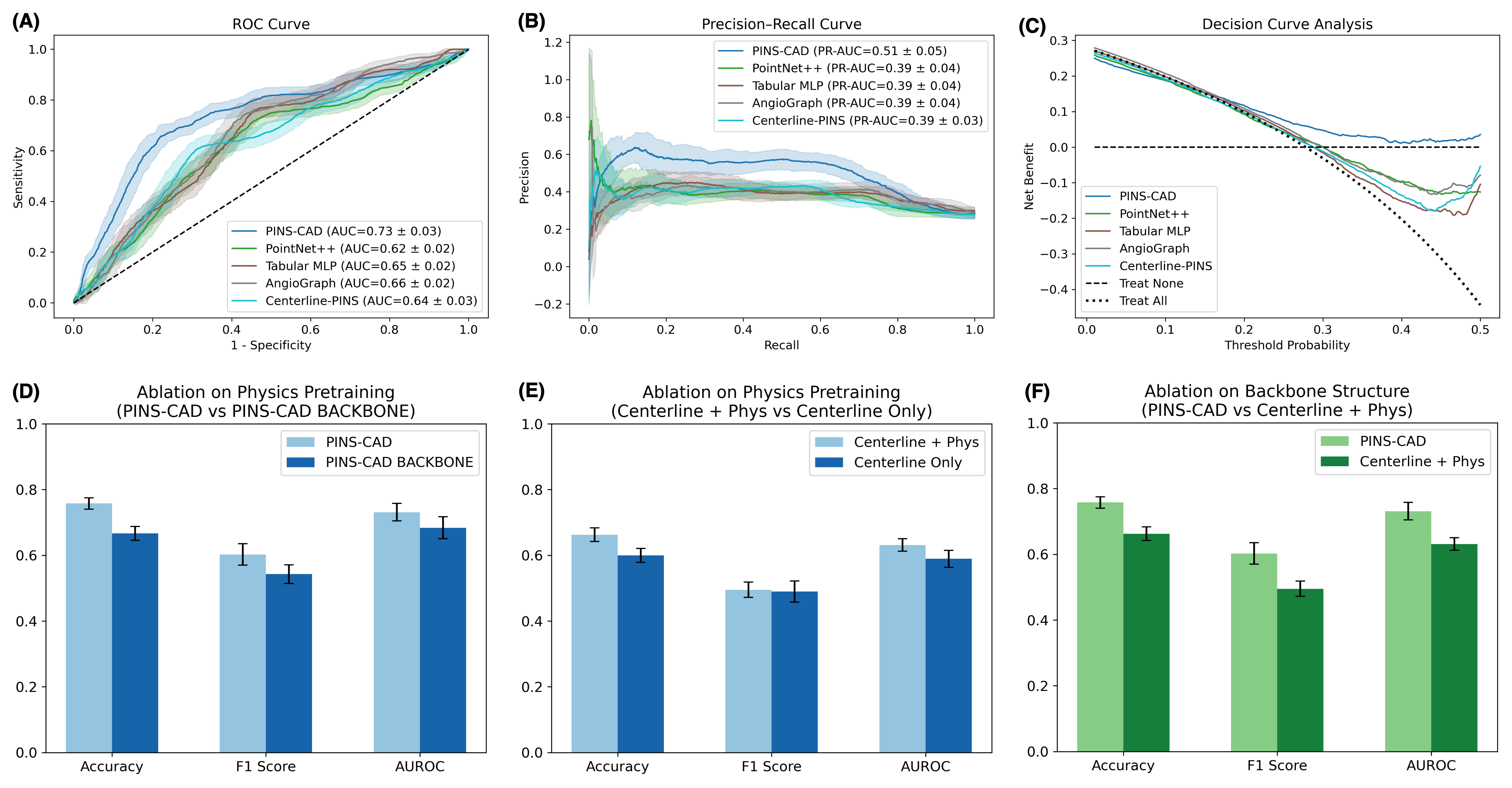}
    \caption{Model performance and ablation analysis of PINS-CAD. \textbf{A}: Receiver operating characteristic (ROC) curves and AUROC values (mean ± standard deviation) for PINS-CAD and four baseline models: PointNet++, Tabular MLP, AngioGraph, and Centerline-PINS.
\textbf{B}: Precision–recall (PR) curves with PR-AUC values for the same models, with shaded areas representing the standard deviation.
\textbf{C}: Decision curve analysis showing net benefit across threshold probabilities, including the ``Treat None'' and ``Treat All'' strategies as references.
\textbf{D-F}: Ablation analyses. All bar plots show mean ± standard deviation. \textbf{D}, Effect of physics-informed pretraining: comparison between PINS-CAD, which uses physics-informed pretraining followed by supervised fine-tuning, and the same backbone trained from scratch without pretraining. \textbf{E}, Effect of physics-informed pretraining under centerline-only input: comparison between a model using physics-informed pretraining and another trained from scratch with supervised learning only. \textbf{F}, Comparison between PINS-CAD and Centerline + Phys, both using physics-informed pretraining but differing in backbone architecture (artery-graph vs. centerline-based), to assess the impact of model structure.}
    \label{fig:comparison}
\end{figure}
We evaluate the proposed PINS-CAD against four representative baselines: PointNet++, Tabular-MLP, AngioGraph, and a centerline-based model with physics-informed pretraining (Centerline-PINS). These models differ in terms of input representation and architectural design, enabling a fair comparison across geometry-based, clinical, and physics-informed learning approaches. PointNet++\cite{qi2017pointnet++} directly takes 3D vessel point clouds as input. Tabular-MLP processes structured clinical features represented as a 35-dimensional vector constructed from 19 tabular clinical variables, including sex, FFR, and DS. The feature vector is then passed through a multilayer perceptron for prediction. AngioGraph\cite{sun2024future} is originally designed for event prediction from 2D coronary angiography by approximating the lesion as a graph, where a subset of lesion pixels are sampled as nodes and their spatial relationships define the edges. To enable a fair comparison across modalities while maintaining consistent vascular representation, we adapted AngioGraph to operate on the 3D digital twin framework by directly using the vascular graph constructed in PINS-CAD as its input, replacing the original 2D lesion graph derived from ICA. Centerline-PINS uses vessel centerlines instead of full 3D point clouds, providing a simplified geometric representation that retains vessel topology but omits boundary-shape information. This baseline allows us to assess the contribution of detailed surface geometry to hemodynamic representation learning. Centerline-PINS is pretrained with the same physics-informed self-supervised strategy as PINS-CAD, followed by supervised fine-tuning. More details about each baseline are provided in the Method section.

Receiver operating characteristic (ROC) and precision–recall (PR) curves are first used to evaluate the performance. As shown in Fig.~\ref{fig:comparison}.A and Fig.~\ref{fig:comparison}.B, PINS-CAD consistently outperforms all baseline methods across multiple evaluation metrics. PINS-CAD achieves the highest area under the ROC curve (AUROC) of 0.73±0.03 and best area under the PR curve (AUPRC) of 0.51±0.05. In comparison, the baseline models show lower prediction ability. Specifically, their AUROC values range from 0.62 (PointNet++) to 0.66 (AngioGraph), representing a 10–15\% relative decrease compared to PINS-CAD. Notably, as illustrated in Fig.~\ref{fig:comparison}.B all baselines yield nearly identical AUPRC values of 0.39, underscoring a limited capacity to predict positive cases under class imbalance. This consistent performance gap across both metrics highlights the superior prediction ability of PINS-CAD.

To evaluate clinical utility, we perform decision curve analysis, which quantifies net benefit~\cite{vickers2016net} as the balance between true and false positives across a range of decision thresholds. In this context, the “treat-all” strategy assumes that all patients are positive and receive intervention, while the “treat-none” strategy assumes no treatment for any patient. These two strategies serve as clinical reference baselines, with useful models expected to yield higher net benefit than both. Decision curve provides insight into the practical value of a model's predictions in clinical settings. As shown in Fig.~\ref{fig:comparison}.C, PINS-CAD achieves the highest net benefit across a broad range of threshold probabilities. In decision curve analysis, the threshold probability represents the clinical decision point at which an intervention is initiated based on the predicted event risk. When the threshold reaches approximately 0.2, PINS-CAD begins to outperform all baseline models as well as the treat-all reference. At thresholds greater than 0.3, PINS-CAD is the only model that maintains a positive net benefit, highlighting its practical value in decision-making scenarios where other methods fail to offer clinical benefit. Although all baseline models yield higher net benefit than the treat-all strategy in the threshold range of 0.3 to 0.5, their net benefit remains negative, indicating no meaningful advantage in guiding treatment decisions.

To investigate the contributions of physics-informed pretraining and input representation, we conduct a series of ablation studies. Fig.\ref{fig:comparison}.D and Fig.\ref{fig:comparison}.E focus on evaluating the impact of physics-informed pretraining, while Fig.~\ref{fig:comparison}.F illustrates the effect of input representation under a fixed training strategy. In Fig.~\ref{fig:comparison}.D, we compare PINS-CAD with its variant trained from scratch without physics-informed pretraining. Across accuracy, AUROC and F1-score, PINS-CAD consistently outperforms its non-pretrained variant, demonstrating the benefit of incorporating domain priors into the learning process. We further assess whether physics-informed pretraining remains beneficial when the framework operates on a simplified representation that uses only the arterial centerline as input. As shown in Fig.~\ref{fig:comparison}.E, even under reduced geometric complexity, the pretrained Centerline-PINS substantially outperforms its non-pretrained variant (Centerline only), confirming that physics-informed self-supervision provides a valuable inductive bias that is independent of input granularity. 

In Fig.~\ref{fig:comparison}.F, we compare Centerline-PINS and PINS-CAD under the same pretraining strategy to isolate the effect of input representation. PINS-CAD consistently outperforms Centerline-PINS across all evaluation metrics, including AUROC, F1-score, and accuracy. This performance gain arises from the difference in feature representation: PINS-CAD first extracts boundary features from the 3D digital twin of the artery using a graph-based encoder applied to the vessel boundary. These boundary-derived features are then integrated into the centerline representation, providing each centerline point with rich contextual information. In contrast, Centerline-PINS relies solely on handcrafted geometric descriptors from the centerline itself, without access to boundary-level structural cues. These results demonstrate that incorporating boundary-derived features significantly enhances the expressiveness of centerline-based models, leading to more accurate predictions.

\subsection*{PINS-CAD enables prediction in DS-FFR discordant lesions}
\begin{figure}
    \centering
    \includegraphics[width=0.95\linewidth]{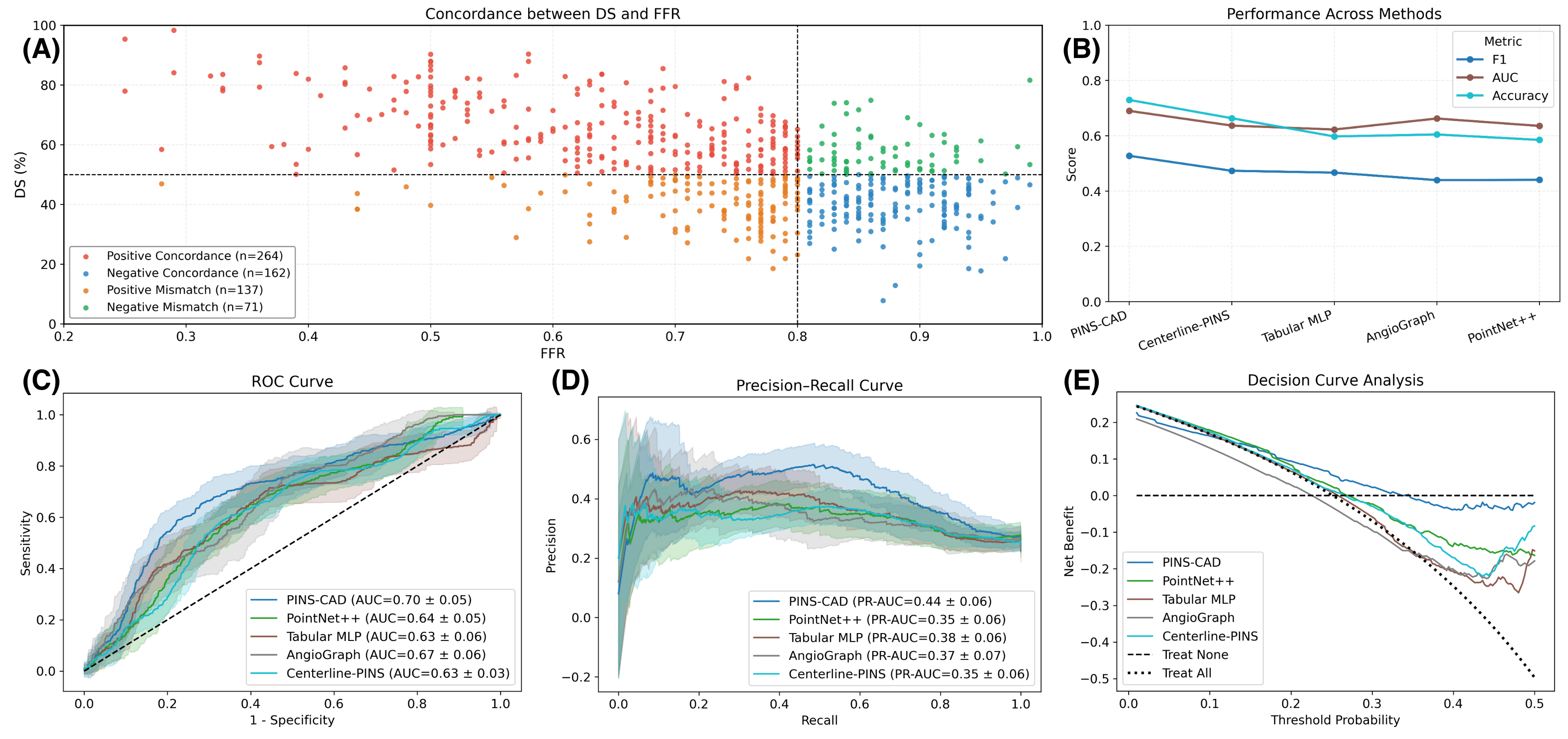}
    \caption{Model performance of PINS-CAD and baseline methods in the contradictory region of FFR and DS. \textbf{A}: Scatter distribution of FFR and DS values across all lesions, illustrating regions of concordance and discordance based on the diagnostic thresholds (FFR = 0.80, DS = 50\%). \textbf{B}: Comparison of overall performance metrics (Accuracy, AUROC, and F1-score) for PINS-CAD and four baseline models: PointNet++, Tabular-MLP, AngioGraph, and Centerline-PINS. \textbf{C}: Receiver operating characteristic (ROC) curves and corresponding AUROC values (mean ± standard deviation) across all methods. \textbf{D}: Precision–recall (PR) curves and PR-AUC values for the same models, with shaded areas indicating the standard deviation.\textbf{E}: Decision curve analysis illustrating net benefit across threshold probabilities, including “Treat None” and “Treat All” strategies as references.}
    \label{fig:contradictory}
\end{figure}
Although FFR and DS are routinely combined to guide coronary revascularization, they frequently provide inconsistent assessments of lesion severity due to their fundamentally different physiological and anatomical foundations~\cite{ciccarelli2018angiography, park2012visual, toth2014evolving}. As shown in Fig.~\ref{fig:contradictory}.A, the scatter distribution of DS and FFR reveals substantial variability between the two indices, with both positively and negatively discordant cases observed. Specifically, 32.8\% of lesions exhibit discordance between the FFR and DS classification thresholds (FFR = 0.80 and DS = 50\%), indicating that one-third of stenoses are interpreted differently depending on whether functional or anatomical assessment is used.
Positive concordance (FFR $\leq$ 0.80, DS $\geq$ 50\%) accounts for 264 lesions (41.6\%), while negative concordance (FFR $>$ 0.80, DS $<$ 50\%) represents 162 lesions (25.6\%). In contrast, 137 lesions (21.6\%) demonstrate positive mismatch (FFR $\leq$ 0.80, DS $<$ 50\%), and 71 lesions (11.2\%) show negative mismatch (FFR $>$ 0.80, DS $\geq$ 50\%).
These discordant subsets represent clinically ambiguous regions where neither anatomical nor physiological assessment alone provides a reliable indicator of ischemic significance.

These discordant regions pose the greatest uncertainty in clinical decision-making, as neither anatomical nor physiological indicators alone provide a reliable basis for intervention. To evaluate whether PINS-CAD maintains discriminative capability within the FFR–DS contradictory region, we specifically analyze model performance in the this region. 
Fig.~\ref{fig:contradictory}.B–E summarize comparative results across five models including the PINS-CAD, Centerline-PINS, Tabular-MLP, AngioGraph, and PointNet++. Although all models show a general decrease in predictive performance compared with the full-cohort results, PINS-CAD maintains the highest overall discrimination.
In terms of overall discrimination as shown in Fig.~\ref{fig:contradictory}.B, PINS-CAD achieves the highest classification performance across AUROC, F1-score, and Accuracy, maintaining predictive ability even when anatomical and physiological cues are inconsistent. Centerline-PINS performs second-best, highlighting the contribution of physics-informed pretraining but also revealing the added benefit of full 3D arterial geometry in PINS-CAD.
Similarly, ROC and PR curves in Fig.~\ref{fig:contradictory}.C–D demonstrate consistent superiority of PINS-CAD over all baseline models, with wider separation margins than observed in the full-cohort analysis. In the corresponding decision curve analysis (Fig.~\ref{fig:contradictory}.E), PINS-CAD surpasses all baseline models as well as the “treat-all” and “treat-none” reference strategies, yielding the highest net clinical benefit across a clinically relevant decision-threshold range of 0.15–0.35. Beyond this interval, the net benefit of all models gradually declines and eventually falls below zero, reflecting the increasing difficulty of reliable decision-making in highly uncertain regions of the FFR–DS space. Nevertheless, PINS-CAD maintains a consistently larger positive margin than all other approaches before this decline, highlighting its ability to extract physiologically meaningful cues even in borderline or contradictory cases. These results demonstrate that physics-informed representation learning provides tangible decision-making value under conditions where traditional anatomical and functional metrics become unreliable.

\subsection*{Learning from anatomically diverse synthetic data improves future cardiovascular events prediction}

To evaluate the impact of anatomical diversity in synthetic digital twins on prediction performance, we compare three model variants: the full PINS-CAD framework pretrained on the large synthetic dataset, a variant pretrained only on real digital twins (SysAblation), and a baseline model trained from scratch without any pretraining (Backbone). We evaluate model performance using accuracy, F1 score, AUROC and AUCPR.
Unlike generic data augmentation, our coronary anatomy-aware strategy generates anatomically realistic and physiologically diverse digital twins by recombining real vessel centerlines and radius profiles, reflecting the variability observed in clinical anatomies. Trained on this anatomically grounded synthetic dataset, PINS-CAD achieves an accuracy of 0.757, an AUROC of 0.724, and an F1 score of 0.599. It outperforms SysAblation, which is trained only on real anatomical data and achieves an accuracy of 0.677, an AUROC of 0.703, and an F1 score of 0.568. The use of synthetic anatomical diversity, generated through structured augmentation of clinical cases, exposes PINS-CAD to a broader distribution of vascular geometries and hemodynamic patterns during pretraining. This exposure facilitates the learning of physiologically consistent and robust features that improve performance on cardiovascular event prediction.
Both pretrained models outperform the baseline trained from scratch, which achieves an accuracy of 0.667, an AUROC of 0.681, and an F1 score of 0.544. These results highlight the effectiveness of physics-informed self-supervision in the absence of CFD labels and emphasize the critical role of anatomical diversity, particularly when generated synthetically, in improving the predictive modeling of coronary artery disease.
\begin{figure}
    \centering
    \includegraphics[width=0.5\linewidth]{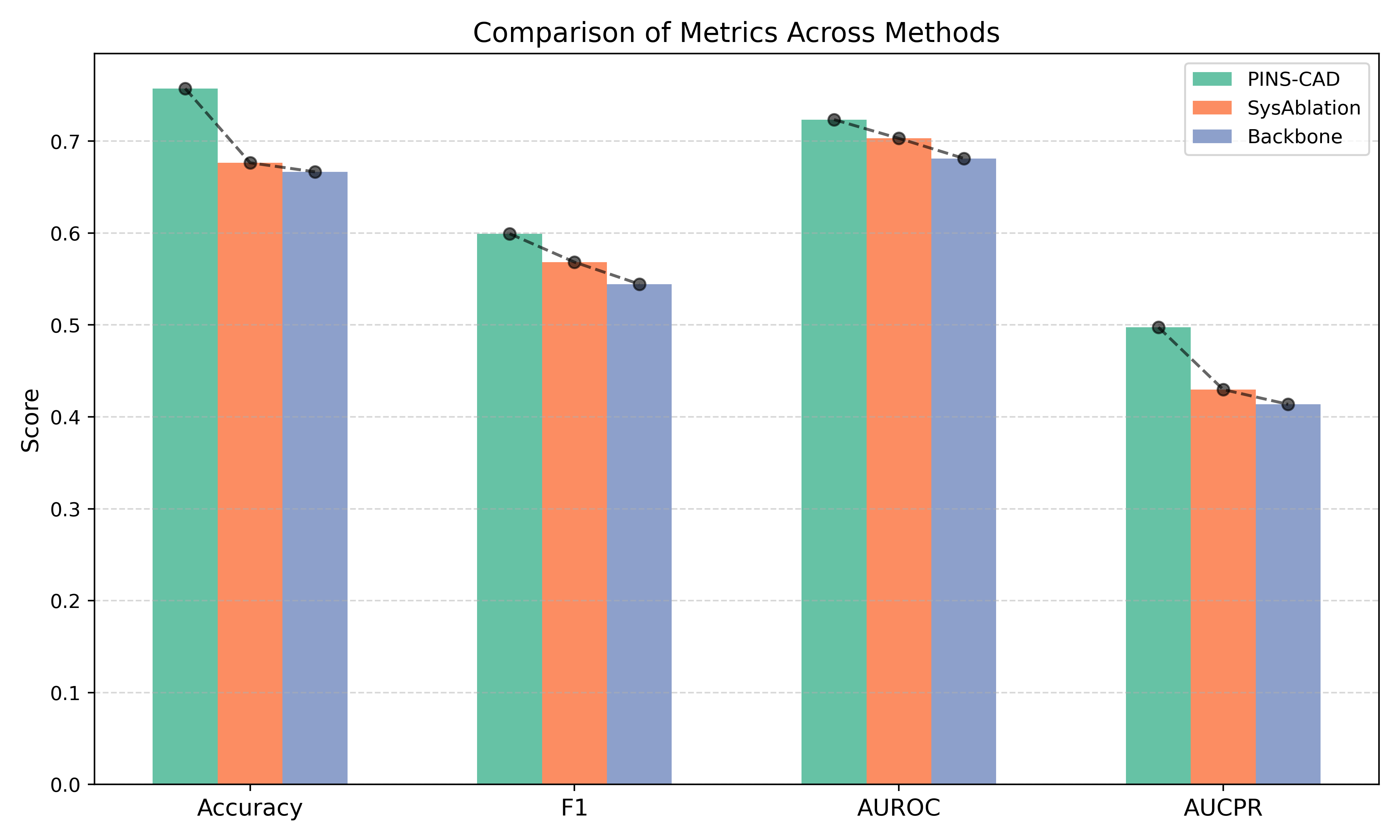}
    \caption{Comparison of model performance with and without synthetic-data pretraining.
PINS-CAD was pretrained on anatomically diverse synthetic digital twins. SysAblation used the same physics-informed pretraining protocol but only on real data. The Backbone model was trained from scratch without any pretraining. Performance was evaluated using accuracy, F1 score, AUROC, and AUPRC.}
    \label{fig:enter-label}
\end{figure}
\subsection*{Predicted pressure patterns and FFR curves enable explainability in PINS-CAD}
\begin{figure}
    \centering
    \includegraphics[width=\textwidth]{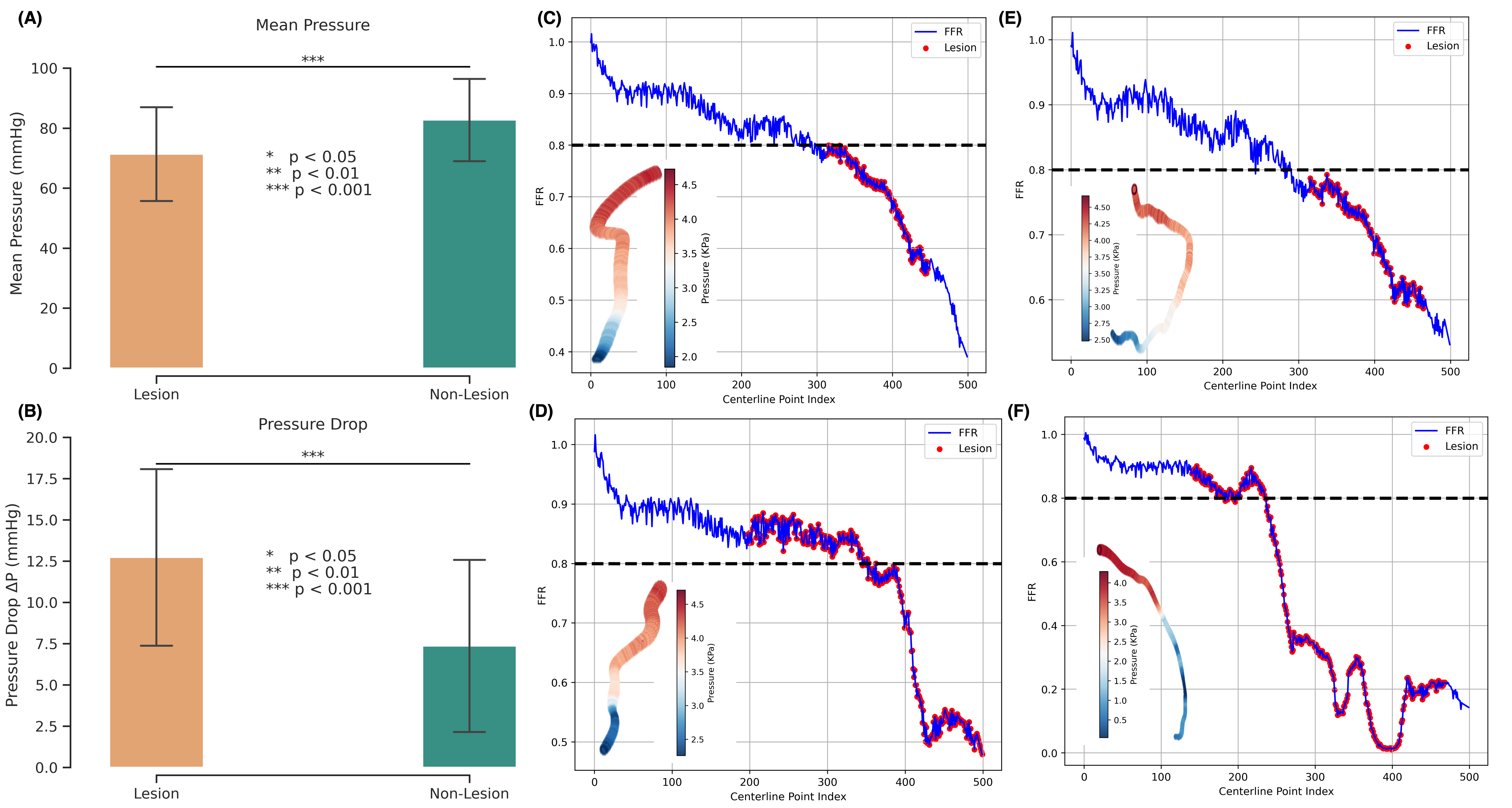}
    \caption{The predicted pressure and FFR from physics-informed self-supervised learning serve as hemodynamic biomarkers, enabling physiological explainability of the model’s predictions. 
    \textbf{A-B}. Comparison of mean pressure and pressure drop between lesion and non-lesion segments. Lesion regions show significantly lower pressure and higher pressure drop, consistent with physiological expectations. Statistical significance is assessed using a two-sided t-test. Error bars denote standard deviation.
    \textbf{C-F}. Four visualization examples of predicted pressure distributions and corresponding FFR curve along the 3D coronary centerline, sampled at 500 evenly spaced points. The left panels in each pair show the pressure distribution mapped onto the artery, while the right panels show the FFR curve computed as the ratio of local pressure to inlet pressure. Red dots indicate lesion regions identified by anatomical labeling. The FFR curves derived from predicted pressure reveal functionally significant drops in flow reserve consistent with lesion severity.}
    \label{fig:visual}
\end{figure}
PINS-CAD enhances model explainability by predicting pressure and FFR curves along the coronary artery, both of which are standard functional parameters in interventional cardiology. Deviations in these patterns are also highly related to lesion severity and coronary diseases. By revealing these hemodynamic signals, PINS-CAD offers intuitive and clinically relevant information that helps clinicians interpret and trust the model’s predictions. PINS-CAD generates continuous pressure predictions and FFR curve along the coronary artery centerline, providing spatially resolved hemodynamic assessment. The pressure is directly predicted by PINS-CAD through the physics-informed self-supervised learning, while the FFR curve is derived by normalizing the predicted pressure relative to its proximal value. 

We first compare the mean pressure and pressure drop ($\Delta P$) between lesion and non-lesion segments across all coronary arteries in the testing data. As shown in Fig.~\ref{fig:visual}.A, lesion regions exhibit significantly lower mean pressure than non-lesion regions, with values of 94.3 ± 8.1 mmHg and 104.7 ± 6.9 mmHg, respectively. Similarly, as shown in Fig.~\ref{fig:visual}.B, the pressure drop is significantly higher in lesion regions than in non-lesion regions, with values of 16.1 ± 5.2 mmHg and 7.8 ± 4.6 mmHg, respectively. Two-sided t-tests indicate that both differences are statistically significant (p < 0.001). These results are consistent with physiological expectations and indicate that the model captures interpretable patterns of pressure reduction across stenotic regions.

We then visualize the predicted pressure and FFR curves along the coronary artery centerline to assess whether PINS-CAD captures localized flow-limiting patterns associated with coronary lesions. To this end, we randomly select four representative samples from the testing set and visualize both the predicted pressure patterns and the corresponding FFR curves (Fig.~\ref{fig:visual}C–F). FFR is calculated as the ratio between the predicted pressure at each point and the proximal reference pressure. Across all four cases, the FFR curves show clear reductions below the clinical threshold of 0.80 within annotated lesion segments, while non-lesion regions remain within normal levels. These spatial patterns are consistent with expected hemodynamic changes in functionally significant stenosis. Although absolute FFR values may not match invasive measurements precisely, the predicted curves successfully capture the declinding trends of pressure loss across lesions. The corresponding 3D pressure visualizations reinforce these findings, highlighting localized pressure drops along the arterial centerline at lesion sites. 
Beyond reproducing FFR-related trends, PINS-CAD provides a richer quantification of lesion-specific hemodynamics by capturing additional flow features such as pressure gradients and velocity distributions. These results demonstrate that PINS-CAD not only captures clinically relevant flow dynamics without supervision from CFD data but also extends hemodynamic assessment beyond conventional FFR, thereby improving both the interpretability and diagnostic value of its predictions.

\subsection*{Evaluating the complementarity of PINS-CAD and clinical tabular data}
\begin{figure}[htbp]
    \centering
    \includegraphics[width=0.8\linewidth]{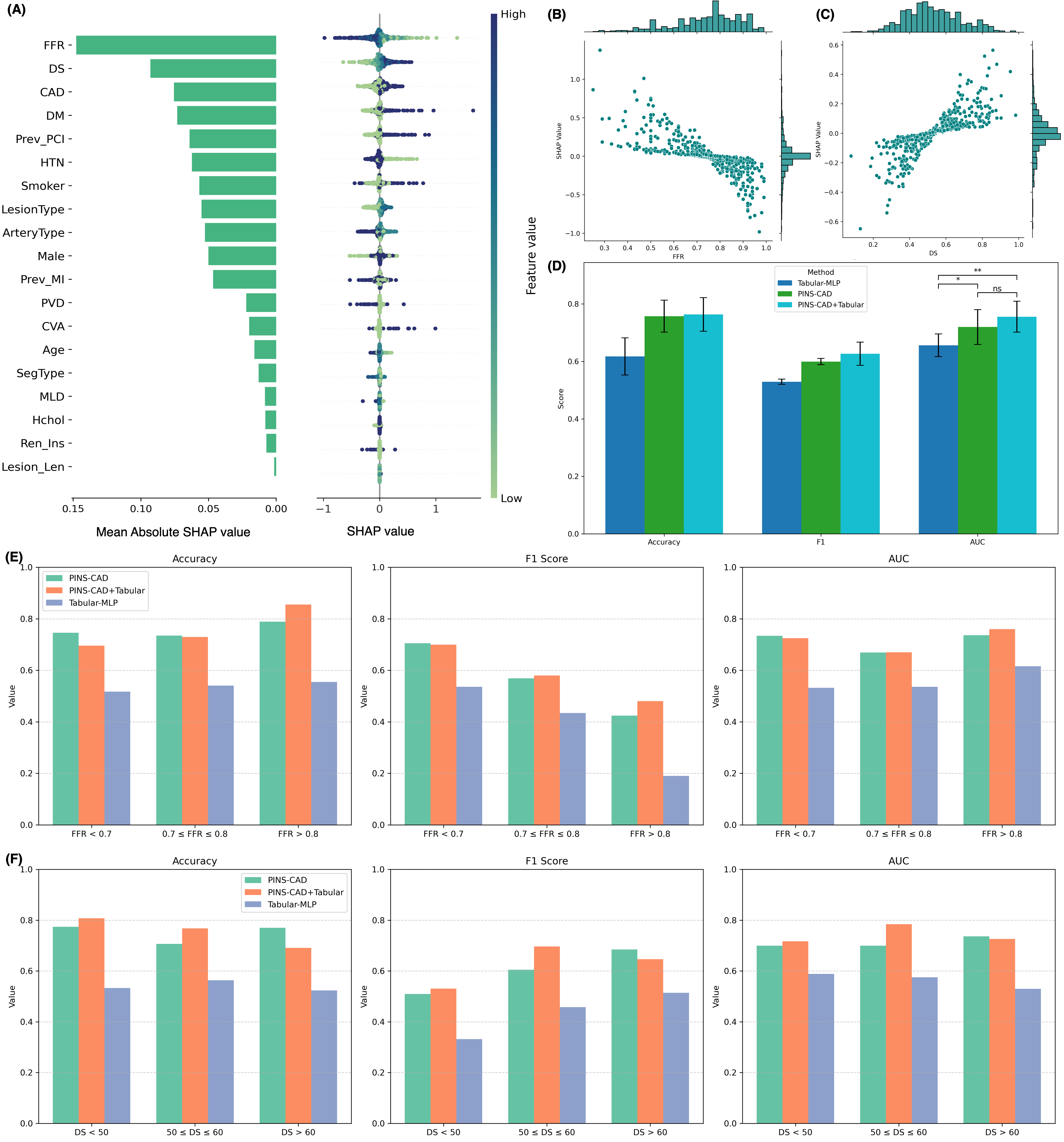}
    \caption{
    Integrating clinical features enhances model interpretability and improves predictive performance.
    \textbf{A.} SHAP summary plot showing the global importance and distribution of clinical tabular features in the Tabular-MLP model. Fractional Flow Reserve (FFR) and Diameter Stenosis (DS) are the most influential variables, followed by established risk factors such as diabetes (DM), hypertension (HTN), and prior PCI. 
    \textbf{B, C.} SHAP dependence plots illustrating the relationship between SHAP values and feature values for FFR (panel B) and DS (panel C). Both plots demonstrate minimal SHAP attribution near the clinical thresholds (FFR $\approx$ 0.8, DS $\approx$ 50\%), indicating low model confidence and increased uncertainty in these borderline regions. 
    \textbf{D.} Performance comparison across the full test set for Tabular-MLP, PINS-CAD, and the fusion model (PINS-CAD+Tabular) in terms of Accuracy, F1 Score, and AUC. Clinical fusion consistently improves performance, particularly in AUC ($p < 0.01$). 
    \textbf{E, F.} Stratified performance by FFR (panel E) and DS (panel F) subgroups. The fusion model outperforms baselines in lower-risk groups (FFR $>$ 0.8, DS $<$ 50\%), while offering marginal or no improvement in higher-risk regions (FFR $<$ 0.7, DS $>$ 60\%).
    }
    \label{fig:tabularanalysis}
\end{figure}

While PINS-CAD effectively integrates geometric and hemodynamic cues, clinical decision-making also relies on patient-level factors such as comorbidities and prior interventions. To quantify what prognostic information is contained within clinical data alone, and to establish a reference for subsequent multimodal fusion, we analyzed the Tabular-MLP model using SHAP (SHapley Additive exPlanations)~\cite{lundberg2017unified}.
SHAP quantifies each feature’s marginal contribution to model predictions, providing interpretable insights into how the model weighs anatomical, functional, and demographic risk factors.
SHAP is a game-theoretic approach that attributes a marginal contribution value to each feature per prediction, enabling interpretability of complex machine learning models. Positive SHAP values indicate a feature’s contribution toward predicting future cardiovascular events, whereas negative values steer predictions toward the absence of events. This analysis provides a baseline understanding of what prognostic cues are embedded in clinical data alone before integrating them with PINS-CAD representations.

As shown in Fig.~\ref{fig:tabularanalysis}.A, FFR and DS are identified as the most influential variables, exhibiting the highest mean absolute SHAP values. Additional relevant features included diabetes (DM), hypertension (HTN), and history of percutaneous coronary intervention (PCI), aligning with known clinical risk factors. This ordering is consistent with clinical guidelines that prioritize both functional and anatomical severity when assessing CAD risk. Beyond global feature importance, Fig.~\ref{fig:tabularanalysis}.B and Fig.~\ref{fig:tabularanalysis}.C describe the distribution of SHAP values across the value ranges of FFR and DS. Low FFR values, near the clinical threshold of 0.8, are consistently associated with positive SHAP scores, suggesting a strong contribution to event prediction. In contrast, higher FFR values shifted model predictions toward the negative class. For DS, the pattern is reversed: high stenosis percentages ($>50\%$) drove the model toward predicting adverse outcomes, while lower values had attenuated influence. These results demonstrate the model’s ability to capture clinically meaningful thresholds and nonlinear interactions among risk factors.

To assess the added value of integrating clinical tabular information, we constructed a fusion model (PINS-CAD+Tabular) by concatenating the deep representations learned from PINS-CAD with handcrafted clinical features. Specifically, the 512-dimensional artery-level embedding generated by the PINS-CAD encoder is concatenated with 34 clinical tabular features and fed into a five-layer multilayer perceptron (MLP) for outcome prediction. We systematically compared the performance of this fusion model with the original PINS-CAD framework and the clinical-only Tabular-MLP baseline across accuracy, F1 score, and AUC on both the full test set and within subgroups stratified by FFR and DS. The subgroups are defined according to established clinical thresholds, where 0.8 was used for FFR and 50\% for DS. These cutoffs align with SHAP findings, which shows minimal attribution around the clinical thresholds, indicating that borderline cases are particularly challenging for the model to predict.

As shown in Fig.~\ref{fig:tabularanalysis}.D, the fusion model achieves the highest AUC of 0.756 (95\% CI: 0.702–0.810) On the full test set, improving upon PINS-CAD by 3.6\% and significantly outperforming Tabular-MLP ($p < 0.01$). Accuracy and F1 score also improve with the addition of clinical features, reaching 0.763 (95\% CI: 0.705–0.822) and 0.627 (95\% CI: 0.586–0.667), respectively. Detailed statistics are provided in Supplementary Table~\ref{tab:Tabular_abl_alldata}. 
On the FFR and DS subgroups, as shown in Fig.~\ref{fig:tabularanalysis}.E and Fig.~\ref{fig:tabularanalysis}.F, both PINS-CAD and the fusion model consistently outperformed the Tabular-MLP across all subgroups. However, the relative gain from adding clinical information varies with lesion severity. In low-risk regions (FFR$>$0.8 or DS$<$50\%), geometric and hemodynamic alterations are minimal, and the morphological cues captured by PINS-CAD are correspondingly weak. These arteries typically present near-normal structure and flow characteristics, making predictions based solely on geometric features more uncertain. In such cases, patient-level clinical factors, such as diabetes, hypertension, or prior PCI provide complementary information that reflects overall cardiovascular condition and helps refine risk estimation. As a result, incorporating clinical variables enhances model discrimination in these subtle cases where anatomical and functional deviations are less pronounced. In contrast, in higher-risk regions (FFR$<$0.7 or DS$>$60\%), geometric and hemodynamic abnormalities are already evident and are effectively captured by PINS-CAD, leaving limited room for further improvement through the inclusion of clinical data. In some instances, the additional tabular information may introduce redundant or weakly relevant signals, slightly reducing overall performance. These results indicate that PINS-CAD alone is sufficient for lesions with clear pathological geometry or flow impairment, whereas the integration of clinical information is particularly beneficial when anatomical and physiological indicators provide less distinct cues.




\section*{Discussion}

This study advances preventive cardiology by introducing PINS-CAD, a physics-informed learning framework that integrates physical fluid dynamics principles with self-supervised learning on personalized 3D coronary artery digital twins to enable early prediction of future cardiovascular events. Guided by hemodynamic principles including the 1D Navier–Stokes and the pressure-drop law, PINS-CAD learns pressure and flow patterns directly from artery geometry without CFD supervision. To enable large-scale physics-informed pretraining, we generate 200,000 synthetic coronary digital twins through anatomy-aware augmentation to improve the robustness and scalability of representation learning. Overall, PINS-CAD demonstrates that coupling physical priors with anatomy-aware augmentation and self-supervised learning can achieve accurate and label-efficient prediction of cardiovascular events, paving the way for more efficient and explainable, evidence-based clinical decision-making in coronary artery disease. 

The superior performance of PINS-CAD suggests that physical priors enable the model to learn representations that reflect underlying coronary physiology. Unlike models trained with purely anatomical objectives~\cite{beetz2023multi}, PINS-CAD employs a physics-informed pretraining strategy that enforces pressure consistency across the coronary arteries through hemodynamic constrains. This inductive bias guides the model to capture stenosis-relevant features and learn latent representations that align with functional surrogates, even without CFD supervision. These hemodynamic abnormalities such as pressure drops and flow disturbances are closely associated with lesion progression and future cardiovascular events, making them physiologically informative for risk prediction. This finding aligns with prior evidence that abnormal pressure gradients and flow resistance are predictive of adverse cardiovascular outcomes.

Anatomical variability in synthetic digital twins strengthens the effectiveness of pretraining in PINS-CAD. Self-supervised learning relies heavily on diverse data, yet obtaining large-scale digital twins is challenging. Our synthetic digital twins are generated through structured geometric augmentation, introducing controlled variations in vessel length, curvature and stenosis severity. This augmented anatomical diversity produces a richer distribution of hemodynamic conditions, exposing the model to a broader spectrum of pressure and flow patterns during pretraining. As a result, the learned representations better reflect physiologically meaningful variations and transfer more effectively to downstream future cardiovascular events prediction. Empirical comparisons show that pretraining using synthetic data yields consistent improvement across all metrics over pretraining using real-world data, indicating that anatomical diversity provides a critical inductive bias for learning robust and transferable representation learning in coronary digital twins.

Moreover, the graph-based backbone enables PINS-CAD to jointly encode geometric features, including diameter, length, and narrowing, along with relational information from anatomically connected segments. This facilitates the modeling of localized flow-relevant interactions along the artery. By stacking multiple message-passing layers, the GNN model also captures longer-range dependencies, allowing it to reason over extended anatomical paths and infer global pressure dynamics across the coronary artery. In comparative experiments, replacing the full artery graph with a simplified centerline representation, which retains coarse geometric paths but omits explicit segment-to-segment interaction modeling, resulted in a 2.9\% relative drop in AUROC. Baselines based on point clouds and tabular features, which lack structured connectivity and capture only local or isolated attributes, exhibited even larger performance declines. These findings underscore the importance of jointly modeling vascular geometry and topological structure to support physiologically meaningful and clinically effective risk prediction.

Combining physics-informed representations from PINS-CAD with clinical variables improves cardiovascular events prediction across the cohort. The PINS-CAD framework captures hemodynamic and geometric features derived from the coronary anatomy, including pressure distribution, flow resistance, and vessel narrowing, which directly reflect artery-level functional significance. In contrast, clinical variables describe systemic risk factors such as age, diabetes, hypertension, and smoking history that influence disease progression beyond anatomy. Integrating these complementary modalities yields more accurate and comprehensive prediction. However, fusion provides limited benefit in borderline cases, particularly for lesions with FFR or DS values near established clinical thresholds, where even traditional clinical assessments offer little discriminative value. Notably, PINS-CAD representations alone outperform tabular clinical models, indicating that physics-informed features offer greater specificity when conventional metrics are equivocal. 

Beyond predictive accuracy, PINS-CAD improves explainability by estimating pressure distributions that reveal physiologically relevant regions within the coronary artery. Most deep learning models, although effective in risk prediction, provide limited insight into underlying mechanisms and rarely incorporate hemodynamic features that directly reflect disease processes. PINS-CAD bridges this gap through a self-supervised pretraining that predicts relative pressure distributions across coronary segments without CFD supervision. These resulting pressure maps consistently highlight stenotic regions with elevated flow resistance and steep pressure gradients. Quantitative analysis shows that predicted pressure values are significantly lower in stenotic segments compared to non-lesion areas. The predicted pressure drops across lesion sites reflect hemodynamic behavior expected from both physical principles and clinical standards, such as the pressure gradients underlying FFR assessments. 
Although the predicted pressure distributions are less precise than those obtained from CFD-supervised models, they still can reflect key hemodynamic patterns, including lesion-associated pressure drops and distal flow resistance. These functional signatures, while coarse in details, provide sufficient physiological context to support risk prediction and yield interpretable representations that align with known determinants of disease progression. Therefore, by embedding physical constraints into representation learning, PINS-CAD yields accurate, interpretable, and clinically trustworthy predictions that support localization of functionally significant regions for downstream evaluation or intervention.

Although this study advances physics-informed neural networks for cardiovascular events prediction, there are several limitations. First, although the model is trained and evaluated using retrospective data collected from multiple centers, the dataset size from each center remains limited, and the whole dataset may not fully reflect the diversity of patient demographics and disease presentations. In addition, while the synthetic digital twins introduce controlled anatomical variation through structured augmentation, they are still generated from a finite set of real cases and may not encompass the full spectrum of coronary morphologies and physiological conditions relevant to flow dynamics. Together, these limitations may constrain the generalizability of the learned representations. Future work should incorporate external validation using larger and more heterogeneous cohorts to evaluate model generalizability. Second, although the pressure estimation during the self-supervised pretraining stage provides informative hemodynamic guidance, it is not as quantitatively accurate as CFD-derived measurements. Due to the scarcity of CFD-labeled data, formal quantitative validation is not conducted. Third, our multimodal fusion strategy relied on simple feature concatenation to combine embeddings from PINS-CAD and clinical information. While this approach yields performance improvements, it may not fully capture the complementary strengths of each modality. Future work should investigate more advanced fusion strategies to better model inter-dependencies and improve predictive accuracy.

\section*{Method}

\subsection*{Images to digital twins pipeline}
The digital twin of each artery was reconstructed from two 2D ICAs. Briefly, the technique utilizes two nearly orthogonal ICA projections. For a given ICA projection, a single frame is extracted at ﬁxed time point, and each ICA image is positioned in 3D space based on acquisition parameters. The coronary artery is segmented \cite{mahendiran2025angiopy} to determine arterial thickness and centerline. Epipolar lines from both views intersect to define 3D points, and integrating thickness data enables the construction of a 3D arterial lumen mesh. Each digital twin is composed of 1,000 cross-sections, with each cross-section defined by 64 3D points, resulting in a total of 64,000 boundary points. The center points of these 1,000 cross-sections define the centerline of coronary artery digital twin.

\subsection*{A\textsuperscript{3}M: Anatomy-aware augmentation module for coronary artery digital twin synthesis}

To enable scalable self-supervised pretraining, we generate synthetic coronary artery digital twins by recombining geometric components from two real patient anatomies. As illustrated in Fig.\ref{fig:framework}.b, each synthetic digital twin is constructed by pairing the centerline of one digital twin with the radius profile of another, thereby decoupling morphology and vessel thickness while preserving anatomical plausibility. Given two real digital twins, denoted as $\mathcal{T}_A = (\mathbf{C}_A, \mathbf{r}_A)$ and $\mathcal{T}_B = (\mathbf{C}_B, \mathbf{r}_B)$, where $\mathbf{C} \in \mathbb{R}^{N \times 3}$ represents the centerline coordinates and $\mathbf{r} \in \mathbb{R}^N$ the corresponding radius profile, a sequence of geometric augmentations are applied to the centerline $\mathbf{C}_A$, including 3D rotation \( \mathbf{R}_\theta \), sinusoidal bending \( \mathcal{B}_\alpha(\cdot) \), and Gaussian smoothing \( \mathcal{G}_\sigma(\cdot) \), to simulate anatomical variability. The augmented centerline is defined as
\begin{equation}
\mathbf{C}_A^{\text{aug}} = \mathcal{G}_\sigma\big( \mathcal{B}_\alpha(\mathbf{R}_\theta \mathbf{C}_A) \big),
\end{equation} where \( \mathbf{R}_\theta \) applies a random 3D Euler-angle rotation, \( \mathcal{B}_\alpha(\cdot) \) introduces low-frequency curvature perturbations to mimic natural vessel tortuosity, and \( \mathcal{G}_\sigma(\cdot) \) performs 1D Gaussian smoothing along the arc length. The resulting centerline $\mathbf{C}_A^{\text{aug}}$ is resampled to a fixed number of $N$ points and normalized to unit scale.

To simulate geometric variability, a Gaussian perturbation $\boldsymbol{\epsilon} \sim \mathcal{N}(0, \sigma^2)$ is applied to the radius profile $\mathbf{r}_B$ to obtain the augmented radius $\mathbf{r}_B^{\mathrm{aug}} = \mathbf{r}_B + \boldsymbol{\epsilon}$. To construct the 3D vessel boundary, each centerline point $\mathbf{c}_i^{\mathrm{aug}} \in \mathbf{C}_A^{\mathrm{aug}}$ is assigned a circular cross-section of radius $r_i^{\mathrm{aug}}$, centered at $\mathbf{c}_i^{\mathrm{aug}}$ and oriented orthogonally to the vessel trajectory. The local tangent vector $\mathbf{t}_i$ is estimated using finite differences between neighboring centerline points. An orthonormal frame $(\mathbf{t}_i, \mathbf{n}_i, \mathbf{u}_i)$ is defined at each location, where $\mathbf{n}_i$ is a unit vector orthogonal to $\mathbf{t}_i$, and $\mathbf{u}_i = \mathbf{t}_i \times \mathbf{n}_i$ denotes the binormal.

Each cross-sectional contour is discretized into $K = 64$ equiangular points. We denote the $j$-th boundary point on the $i$-th cross-section as $\mathbf{q}_{i,j} \in \mathbb{R}^3$, defined by:
\begin{equation}
\mathbf{q}_{i,j} = \mathbf{c}_i^{\mathrm{aug}} + r_i^{\mathrm{aug}} \left( \cos \theta_j \, \mathbf{n}_i + \sin \theta_j \, \mathbf{u}_i \right), \quad \theta_j = \frac{2\pi j}{K}, \quad j = 0, 1, \dots, K{-}1.
\end{equation}

This sweeping process results in a continuous, anatomically plausible surface that captures variability in both vessel trajectory and thickness.

\subsection*{Graph modeling of 3D digital twins}

To effectively model digital twins, we represent them as graphs
leveraging their ability to capture complex geometric structures and spatial relationships between anatomical points. The graph $G = (V, E, \mathbf{F})$ is constructed in three key steps: node definition, edge construction and node feature encoding.

\subsubsection*{Graph node definition}
Each node \( v_i \in V \) represents a 3D point on the boundary of the digital twin, defined by its spatial coordinates \( \mathbf{p}_i = (x_i, y_i, z_i) \). These points are sampled from uniformly spaced anatomical sections along the vessel centerline, with a fixed number of boundary points extracted per section in a consistent angular order. The resulting set of nodes encodes the overall luminal geometry of the vessel.

\subsubsection*{Graph edge construction}
To encode both the local geometric structure within each cross-section and the continuity along the vessel axis, we construct the edge set \( E \) by combining intra- and inter-sectional connections. Within each section, edges are added between each point and its two neighbors based on their ordering, forming a circular topology that preserves circumferential relationships. To enforce longitudinal coherence, each point is further connected to its three nearest neighbors in the preceding and following sections, identified using a Euclidean \( k=3 \) nearest neighbors (KNN). As a result, each node is connected to a total of eight neighboring nodes: two within the same section and six from adjacent sections. This results in a spatially structured graph that reflects both local surface geometry and global vessel continuity.

\subsubsection*{Graph feature encoding}
Each node \( v_i \) is associated with a feature vector \( \mathbf{f}_i = \left[ \mathbf{p}_i, a_i, r_i \right] \in \mathbb{R}^5 \), where \( \mathbf{p}_i = (x_i, y_i, z_i) \) denotes the 3D coordinates of the point, \( a_i \in \mathbb{R} \) is the local luminal cross-sectional area, and \( r_i \in \mathbb{R} \) is the shortest Euclidean distance from the node to the vessel centerline. These features jointly capture geometric and anatomical information necessary for downstream learning tasks.

\subsection*{Graph neural network backbone for representation learning}

Both local geometric features and a global understanding of hemodynamic patterns are essential for predictive modeling of coronary artery digital twins. Flow rate $Q$ and pressure $P$ are key hemodynamic parameters, typically obtained via CFD simulations, which are computationally expensive and time-consuming. To address this, we propose a model $\mathcal{M}(G)$ that directly predicts $Q\in\mathbb{R}^{1 \times N}$ and $P\in\mathbb{R}^{1 \times N}$ at $N$ discrete points along the vessel centerline $C$ of the input graph $G$, optimizing physical losses without labeled data. The model first employs a hierarchical GNN-based encoder to perform message passing over the vascular graph, extracting node-level geometric features. These features are then aggregated onto the centerline via a centerline aggregation module. PI-SSL is then applied to the fused centerline features to supervise the prediction of $P$ and $Q$. The learned centerline-level latent representations can further be fine-tuned for downstream tasks.

\subsubsection*{Hierarchical GNN encoder architecture}
A hierarchical GNN-based encoder is employed to extract multi-level geometric representations from the vascular graph. The encoder consists of three sequential GNN blocks, each comprising four Graph Convolutional Network (GCN) layers~\cite{kipf2016semi} for message passing and local feature aggregation. Between consecutive blocks, a Top-K pooling layer\cite{knyazev2019understanding,gao2019graph,cangea2018towards} with pooling ratio $PR=0.5$ is applied to reduce the number of nodes, retaining the most informative ones and enabling hierarchical abstraction. As the network deepens, the dimensionality of node features is progressively increased from $d$ to $3d$, allowing the encoder to learn increasingly expressive representations of the vascular structure.

\subsubsection*{Centerline Aggregation}
To predict $Q$ and $P$ at each point on $C$, we introduce the Centerline Aggregation (CA) module, which constructs centerline point features by aggregating information from nearby graph nodes. Given graph node coordinates $\mathbf{P} \in \mathbb{R}^{|V| \times 3}$, graph node features $\mathbf{F}\in\mathbb{R}^{|V| \times D}$, where $D \in \{d, 2d, 3d\}$ denotes the dimension of node embeddings produced by the GNN backbone, centerline point coordinates $\mathbf{CP}\in\mathbb{R}^{|N| \times 3}$, and the number of nearest neighbors $K_{ca}$ to consider for each centerline point, the goal is to generate a structured centerline feature $\mathbf{H} \in \mathbb{R}^{N \times K_{ca} \times D}$ that integrates spatial and geometric information from the coronary artery graph.
For each centerline point $\mathbf{CP}_i$, $i \in \{1, \ldots, N\}$,
CA identifies its $K_{ca}$ nearest graph nodes using Euclidean distance and extracts their corresponding node features $\mathbf{F_j}\in\mathbb{R}^{1 \times D}$ and node coordinates $\mathbf{P_j}\in\mathbb{R}^{1 \times 3}$, $j \in \{1, \ldots, K_{ca}\}$. To capture the local geometric structure, the relative coordinates of neighbors with respect to the $\mathbf{CP}_i$ are computed. These neighboring features and relative coordinates are then concatenated to form the centerline feature representation $\mathbf{H}_i\in\mathbb{R}^{K_{ca} \times (D+3)}$ for $\mathbf{CP}_i$. This process is repeated for all centerline points, resulting in the complete centerline feature representation $\mathbf{H}\in\mathbb{R}^{N\times K_{ca} \times (D+3)}$, which is further processed through a convolutional layer to yield $\mathbf{H} \in \mathbb{R}^{N \times K_{ca} \times D}$. To refine the centerline representation, a mean aggregation operation is applied across the $K_{ca}$ neighboring points, producing the final centerline representation $\mathbf{H} \in \mathbb{R}^{N \times D}$. The centerline features from different levels are concatenated and further processed through a multi-layer perceptron (MLP) to refine it into $\mathbf{H} \in \mathbb{R}^{N \times 3d}$.

\subsubsection*{Physics-informed pretraining and downstream fine-tuning}

An MLP consisting of four linear layers is used to progressively reduce the feature dimension from $3d$ to 2, enabling the prediction of $Q$ and $P$ at each centerline point. These predictions are guided by physics-informed loss functions derived from fluid dynamics, allowing self-supervised learning without reliance on CFD-labeled data.

For downstream fine-tuning, the learned centerline features $\mathbf{H} \in \mathbb{R}^{N \times 3d}$ are globally pooled and passed through a separate MLP to predict clinical events. The encoder is frozen during this stage, and the classifier is trained using a cross-entropy loss.

\subsection*{Physics-informed self-supervised pretraining loss}
PINS-CAD incorporates Navier-Stokes-based residual loss and pressure drop loss as fundamental physical constraints to guide self-supervised learning for $Q$ and $P$ prediction without CFD data. These constraints ensure that the model captures intrinsic relationship between flow dynamics and vessel geometry.

\subsubsection*{Navier-Stokes-based residual loss} Cardiovascular blood flow follows the Navier-Stokes (NS) equations, which govern the blood fluid dynamics~\cite{gu2024physics}. While solving 3D NS equations provides detailed hemodynamic insights, it is computationally expensive and impractical for data-driven frameworks. Instead, we adopt a 1D reduced-order NS model, which provides an efficient approximation of blood flow along the arterial centerline while preserving critical hemodynamic properties~\cite{formaggia2003one,hu2023novel}.
For steady-state, incompressible blood flow, the 1D NS equations governing mass and momentum conservation for modeling $Q(x)$ and $P(x)$ are given by:
\begin{equation}
\begin{aligned}
    \frac{\partial Q(x)}{\partial x} &= 0, \quad
    \frac{\partial}{\partial x} \left( \frac{Q(x)^2}{A(x)} \right)
    + \frac{A(x)}{\rho} \frac{\partial P(x)}{\partial x}
    - \frac{2 (\zeta + 2) \mu \pi}{\rho} \frac{Q(x)}{A(x)} = 0,
\end{aligned}
\label{NS}
\end{equation}
where $A(x)$ is the cross-sectional area, $x$ is the axial coordinate along the centerline. The blood density $\rho=1.05g/cm^3$, velocity profile factor $\zeta=4.31$ and viscosity $\mu=0.035 dyne/cm^2$ are consistent with standard physiological conditions~\cite{muller2021impact}. 
To enforce model adherence to Eq.~\ref{NS}, we define the residual loss at each centerline point $x$ as:
\begin{equation}\label{lossresidual}
    {Residual}(x) = \frac{\partial}{\partial x} \left( \frac{Q(x)^2}{A(x)} \right) + \frac{A(x)}{\rho} \frac{\partial P(x)}{\partial x} - \frac{2 (\zeta + 2) \mu \pi}{\rho} \frac{Q(x)}{A(x)}+\frac{\partial Q(x)}{\partial x},
\end{equation} where $Q(x)$ and $P(x)$ are self-supervised predictions. The residual loss is calculated as the mean squared residual over all $N$ points along the centerline:
\begin{equation}
    \mathcal{L}_{\text{residual}} = \frac{1}{N} \sum_{i=1}^{N} \left( \text{Residual}(x_i) \right)^2.
\end{equation}
Minimizing $\mathcal{L}_{\text{residual}}$ ensures the predicted flow rate and pressure fields satisfy the NS equation, enforcing physical consistency in the learning process.


\subsubsection*{Pressure drop loss} 

The pressure drop across a coronary artery is a crucial hemodynamic metric, influenced by lesions and viscous losses. Lesions in the artery cause significant localized pressure drops. Additionally, pressure also decreases naturally in healthy segments due to viscous losses and energy dissipation. To model the overall hemodynamic state, we compute the total pressure drop along the entire centerline, capturing both pathological (lesion-induced) and physiological (healthy segment) contributions. 

The pressure drop in a healthy segment $\Delta P_h$ is computed as~\cite{fossan2018uncertainty}:
\begin{align}\label{droph}
    \Delta P_h &= Q\sum_{i=1}^{N} \frac{2 (\zeta + 2) \pi \mu}{A(x_i)^2} \Delta x + \frac{\rho}{2} \left( \frac{1}{A_{\text{out}}^2} - \frac{1}{A_{\text{in}}^2} \right) Q^2,
\end{align}
where $Q$ is the average flow rate over the segment, $x_i$ is the position of the i-th centerline point, $\Delta x$ is the spacing between two consecutive points in the discrete representation of the vessel, $A_{\text{in}}$ and $A_{\text{out}}$ represent the cross-sectional areas at the inlet (start) and outlet (end) points of the health segment, respectively.
Similarly, the pressure drop in a lesion segment $\Delta P_s$ can be computed by Eq.~\ref{drops}~\cite{fossan2021machine,muller2021impact,seeley1976effect}, 
\begin{align}\label{drops}
    \Delta P_s &= Q \sum_{i=1}^{N} \frac{8 \pi \mu}{A(x_i)^2} \Delta x + \frac{K_t \rho}{2 A_0^2} \left( \frac{A_0}{A_s} - 1 \right)^2 Q^2, 
\end{align}
where the empirical coefficient $K_t$ is set as 1.52~\cite{seeley1976effect}, $A_{0}$ is the average cross-sectional area at the inlet and outlet, $A_{s}$ represents the minimum cross-sectional area in the lesion segment. Eqs.~\ref{droph} and~\ref{drops} define the theoretical pressure drop $\Delta P_{\mathrm{phys}} = \Delta P_{h} + \Delta P_{s}$, which can be directly computed from predicted flow rate and vessel geometry.

To ensure the model's predicted pressure align with these physical constraints, we supervise them from two complementary perspectives including the global pressure loss and local pressure loss. The global loss enforces consistency between the predicted and theoretical pressure drop across the entire artery, while the local loss captures spatial variations in resistance, promoting regional accuracy.

We compute the predicted global pressure drop as $\Delta P_{\mathrm{pred}} = P_{\mathrm{in}} - P_{\mathrm{out}}$, where $P_{\mathrm{in}}$ and $P_{\mathrm{out}}$ are the average pressures at the first and last $k$ points of the centerline. The global consistency loss is defined as:
\begin{equation}
\mathcal{L}_{\text{global}} = \left[ \log \left( 1 + \left| \frac{ \Delta P_{\mathrm{pred}} - \Delta P_{\mathrm{phys}} }{ \Delta P_{\mathrm{phys}} + \varepsilon } \right| \right) \right]^2,
\end{equation}
where $\varepsilon$ is a small constant to prevent division by zero. The log-transformed relative error formulation helps mitigate the influence of outliers during gradient propagation. $\Delta P_{\mathrm{phys}}$ denotes the theoretical pressure drop computed over the entire artery based on predicted flow and vessel geometry, incorporating both healthy and lesion segments as described in Eq.~\ref{droph} and Eq.~\ref{drops}.

To further enhance spatial sensitivity, we introduce a sliding window–based local pressure drop loss. While the global loss enforces overall consistency, it may overlook localized discrepancies—especially in vessels with subtle or diffuse stenoses. Therefore, we divide the artery into a series of overlapping windows and evaluate pressure drop consistency within each segment independently. Within each window, both the predicted and theoretical pressure drops are computed using the same formulations as in the global loss, but restricted to the window boundaries.

For the $m$-th window, the predicted and theoretical pressure drops are denoted by $\Delta P^{(m)}_{\mathrm{pred}}$ and $\Delta P^{(m)}_{\mathrm{phys}}$, respectively. Similar to the global loss, the local loss is defined as:
\begin{equation}
\mathcal{L}_{\text{local}} = \frac{1}{M} \sum_{m=1}^{M} \left[ \log \left( 1 + \left| \frac{ \Delta P^{(m)}_{\mathrm{pred}} - \Delta P^{(m)}_{\mathrm{phys}} }{ \Delta P^{(m)}_{\mathrm{phys}} + \varepsilon } \right| \right) \right]^2,
\end{equation}
where $M$ is the number of sliding windows, and $\varepsilon$ is a small constant as before. This localized supervision encourages the model to learn pressure-flow relationships with finer granularity along the vessel.




\bibliography{ref}






\section*{Supplementary information (optional)}
\subsection*{Image to digital twins}
\begin{figure}
    \centering
    \includegraphics[width=0.9\linewidth]{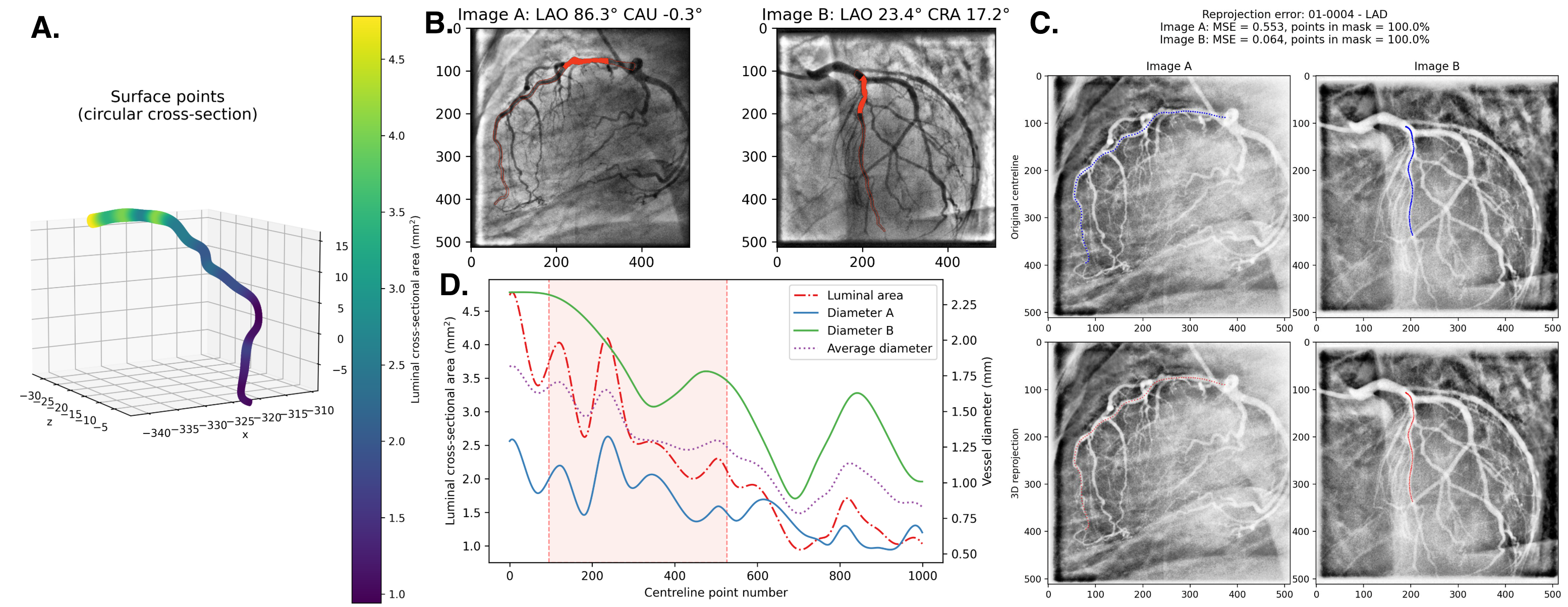}
\caption{
\textbf{3D reconstruction of a coronary artery digital twin from two 2D ICA images.}
\textbf{(A)} Example of a reconstructed 3D coronary digital twins with circular cross-sections, where color encodes the local luminal area.
\textbf{(B)} Corresponding paired angiographic projections (Image A: LAO 86.3°/CAU –0.3°, Image B: LAO 23.4°/CRA 17.2°) used for 3D reconstruction. The extracted vessel centerlines are overlaid in red.
\textbf{(C)} Comparison of original 2D projections (top) and the reprojections of the reconstructed 3D centerline (bottom), demonstrating subpixel reprojection accuracy.
\textbf{(D)} Quantitative luminal cross-sectional area and diameter profiles along the vessel centerline derived from the reconstructed 3D geometry. 
These profiles allow localized assessment of lesion severity and continuity of the vascular structure.
}    
\label{fig:digital_sample}
\end{figure}
Fig.~\ref{fig:digital_sample}.A illustrates the visualization of a reconstructed three-dimensional (3D) coronary artery digital twin, where color encodes the local luminal area along the vessel. The resulting model provides a continuous representation of the coronary lumen, preserving curvature and diameter variation along the centerline. The 3D reconstruction is derived from two angiographic projections, as shown in Fig.~\ref{fig:digital_sample}.B, acquired from nearly orthogonal viewing angles (LAO–CAU and LAO–CRA), which provide complementary depth information. Vessel centerlines are extracted from each 2D projection and fused through epipolar triangulation to recover the 3D spatial trajectory and local radius profile of the artery. Reconstruction accuracy is evaluated by reprojecting the reconstructed 3D centerline onto the original image planes, as shown in Fig.~\ref{fig:digital_sample}.C. The low mean squared error (MSE) between the reprojected and annotated centerlines indicates high geometric consistency between the 2D projections and the 3D reconstruction. Finally, Fig.~\ref{fig:digital_sample}.D presents the derived profiles of luminal cross-sectional area and diameter along the vessel centerline, demonstrating the model’s ability to capture continuous spatial variations in vessel caliber and lesion severity.

\begin{table}[htbp]
\centering
\caption{Performance metrics (mean and 95\% confidence intervals) for each method.}
\label{tab:Tabular_abl_alldata}
\begin{tabular}{llccc}
\toprule
\textbf{Metric} & \textbf{Method} & \textbf{Mean} & \textbf{95\% CI Lower} & \textbf{95\% CI Upper} \\
\midrule
\multirow{3}{*}{Accuracy} 
  & Tabular-MLP         & 0.617 & 0.553 & 0.682 \\
  & PINS-CAD            & 0.757 & 0.702 & 0.813 \\
  & PINS-CAD+Tabular    & 0.763 & 0.705 & 0.822 \\ \hline
\addlinespace
\multirow{3}{*}{F1 Score} 
  & Tabular-MLP         & 0.529 & 0.520 & 0.538 \\
  & PINS-CAD            & 0.600 & 0.589 & 0.610 \\
  & PINS-CAD+Tabular    & 0.627 & 0.586 & 0.667 \\ \hline
\addlinespace
\multirow{3}{*}{AUC} 
  & Tabular-MLP         & 0.656 & 0.617 & 0.696 \\
  & PINS-CAD            & 0.720 & 0.659 & 0.780 \\
  & PINS-CAD+Tabular    & 0.756 & 0.702 & 0.810 \\
\bottomrule
\end{tabular}
\end{table}

\begin{table}[htbp]
\centering
\caption{Performance comparison across FFR subgroups.}
\label{tab:auc_ffr_strata}
\begin{tabular}{llccc}
\toprule
\textbf{FFR Group} & \textbf{Method} & \textbf{Accuracy} & \textbf{F1 Score} & \textbf{AUC} \\
\midrule
\multirow{3}{*}{FFR < 0.7} 
    & Tabular-MLP      & 0.517 & 0.536 & 0.532 \\
    & PINS-CAD         & 0.745 & 0.705 & 0.734 \\
    & PINS-CAD+Tabular & 0.696 & 0.699 & 0.725 \\ \hline
\addlinespace
\multirow{3}{*}{$0.7 \leq$ FFR $\leq 0.8$} 
    & Tabular-MLP      & 0.541 & 0.434 & 0.536 \\
    & PINS-CAD         & 0.735 & 0.569 & 0.669 \\
    & PINS-CAD+Tabular & 0.729 & 0.579 & 0.670 \\ \hline
\addlinespace
\multirow{3}{*}{FFR > 0.8} 
    & Tabular-MLP      & 0.555 & 0.190 & 0.615 \\
    & PINS-CAD         & 0.789 & 0.424 & 0.736 \\
    & PINS-CAD+Tabular & 0.856 & 0.480 & 0.759 \\
\bottomrule
\end{tabular}
\end{table}

\begin{table}[htbp]
\centering
\caption{Performance comparison across DS subgroups.}
\label{tab:auc_ds_strata}
\begin{tabular}{llccc}
\toprule
\textbf{DS Group} & \textbf{Method} & \textbf{Accuracy} & \textbf{F1 Score} & \textbf{AUC} \\
\midrule
\multirow{3}{*}{DS < 50\%} 
    & Tabular-MLP      & 0.533 & 0.332 & 0.588 \\
    & PINS-CAD         & 0.774 & 0.509 & 0.699 \\
    & PINS-CAD+Tabular & 0.808 & 0.531 & 0.717 \\ \hline
\addlinespace
\multirow{3}{*}{50\% $\leq$ DS $\leq$ 60\%} 
    & Tabular-MLP      & 0.563 & 0.458 & 0.575 \\
    & PINS-CAD         & 0.707 & 0.605 & 0.699 \\
    & PINS-CAD+Tabular & 0.767 & 0.697 & 0.784 \\ \hline
\addlinespace
\multirow{3}{*}{DS > 60\%} 
    & Tabular-MLP      & 0.524 & 0.514 & 0.529 \\
    & PINS-CAD         & 0.770 & 0.685 & 0.737 \\
    & PINS-CAD+Tabular & 0.691 & 0.647 & 0.726 \\
\bottomrule
\end{tabular}
\end{table}





\begin{table}[htbp]
\caption{Summary of clinical and angiographic variables with data types, binary distributions, and statistics for numeric variables.}
\centering
\small
\begin{tabular}{lllll}
\toprule
\textbf{Abbreviation} & \textbf{Full Name} & \textbf{Data Type} & \textbf{Binary Distribution} & \textbf{Mean ± SD} \\
\midrule
Lesion\_Length & Lesion Length & Numeric & --- & 13.94 ± 24.74 \\
MLD & Minimal Luminal Diameter & Numeric & --- & 1.30 ± 0.66 \\
DS & Diameter Stenosis (\%) & Numeric & --- & 53.01 ± 15.14 \\
HTN & Hypertension & Binary & 76.94\% Yes / 23.06\% No & --- \\
Hchol & Hypercholesterolaemia & Numeric & --- & 1.68 ± 25.09 \\
DM\_Overall & Diabetes Mellitus (any type) & Binary & 25.56\% Yes / 74.44\% No & --- \\
Ren\_Ins & Renal Insufficiency & Binary & 3.01\% Yes / 96.99\% No & --- \\
PVD & Peripheral Vascular Disease & Binary & 9.65\% Yes / 90.35\% No & --- \\
CVA & Cerebrovascular Accident & Binary & 3.13\% Yes / 96.87\% No & --- \\
Prev\_MI & Previous Myocardial Infarction & Binary & 38.66\% Yes / 61.34\% No & --- \\
Prev\_PCI & Previous Percutaneous Coronary Intervention & Binary & 18.19\% Yes / 81.81\% No & --- \\
Age & Age & Numeric & --- & 64.24 ± 9.93 \\
Male & Male Sex & Binary & 74.81\% Yes / 25.19\% No & --- \\
CAD & Known Coronary Artery Disease & Binary & 47.24\% Yes / 52.76\% No & --- \\
Smoker & Smoker & Binary & 20.55\% Yes / 79.45\% No & --- \\
FFR & Fractional Flow Reserve & Numeric & --- & 0.74 ± 0.15 \\
Lesion\_Type & Lesion Type & Numeric & --- & 2.14 ± 0.95 \\
artery & Artery Type & Categorical & --- & --- \\
SegType & Segment Type & Categorical & --- & --- \\
\bottomrule
\end{tabular}
\label{tab:clinical_summary}
\end{table}

\begin{table}
\centering
\caption{Performance comparison across methods.}
\begin{tabular}{lcccccc}
\toprule
\textbf{Method} & \textbf{Accuracy} & \textbf{F1} & \textbf{AUROC} & \textbf{AUCPR}  \\
\midrule
PINS-CAD     & 0.7574 & 0.5993 & 0.7235 & 0.4975 \\
SysAblation  & 0.6765 & 0.5684 & 0.7031 & 0.4297  \\
Backbone     & 0.6667 & 0.5445 & 0.6813 & 0.4136 \\
\bottomrule
\end{tabular}
\label{tab:method_comparison}
\end{table}

\end{document}